\documentclass[manuscript,screen,nonacm]{acmart}

\AtBeginDocument{%
  }

\setcopyright{none}

\usepackage{comment}
\usepackage{float}
\usepackage{listings}
\usepackage{graphicx}
\usepackage{booktabs}
\begin{document}

\title[Large Language Models for Mental Health Diagnostic Assessments]{Exploring The Potential of Large Language Models for Assisting with Mental Health Diagnostic Assessments\\\textit{The Depression and Anxiety Case}}

\author{Kaushik Roy}
\email{kaushikr@email.sc.edu}
\affiliation{%
  \institution{Artificial Intelligence Institute University of South Carolina}
  \country{USA}
}
\author{Harshul Surana}
\email{harshul19@iiserb.ac.in}
\affiliation{%
  \institution{Indian Institute of Research and Science, Bhopal}
  \country{India}
}
\author{Darssan Eswaramoorthi}
\email{ledarssan@gmail.com}
\affiliation{%
  \institution{Artificial Intelligence Institute University of South Carolina}
  \country{USA}
}
\author{Yuxin Zi}
\email{yzi@email.sc.edu}
\affiliation{%
  \institution{Artificial Intelligence Institute University of South Carolina}
  \country{USA}
}

\author{Vedant Palit}
\email{ledarssan@gmail.com}
\affiliation{%
  \institution{Indian Institute of Technology, Kharagpur}
  \country{India}
}
\author{Ritvik Garimella}
\email{ritvik916@gmail.com}
\affiliation{%
  \institution{Artificial Intelligence Institute University of South Carolina}
  \country{USA}
}

\author{Amit Sheth}
\email{amit@sc.edu}
\affiliation{%
  \institution{Artificial Intelligence Institute University of South Carolina}
  \country{USA}
}

\renewcommand{\shortauthors}{Trovato et al.}
\acmArticleType{Review}

\maketitle

\section*{abstract}
Large language models (LLMs) are increasingly attracting the attention of healthcare professionals for their potential to assist in diagnostic assessments, which could alleviate the strain on the healthcare system caused by a high patient load and a shortage of providers. For LLMs to be effective in supporting diagnostic assessments, it is essential that they closely replicate the standard diagnostic procedures used by clinicians. In this paper, we specifically examine the diagnostic assessment processes described in the Patient Health Questionnaire-9 (PHQ-9) for major depressive disorder (MDD) and the Generalized Anxiety Disorder-7 (GAD-7) questionnaire for generalized anxiety disorder (GAD). We investigate various prompting and fine-tuning techniques to guide both proprietary and open-source LLMs in adhering to these processes, and we evaluate the agreement between LLM-generated diagnostic outcomes and expert-validated ground truth. For fine-tuning, we utilize the Mentalllama and Llama models, while for prompting, we experiment with proprietary models like GPT-3.5 and GPT-4o, as well as open-source models such as llama-3.1-8b and mixtral-8x7b.

\paragraph*{\textbf{Software Availability}}
We make all software artifacts available at this \href{https://github.com/kauroy1994/Large-Language-Models-for-Assisting-with-Mental-Health-Diagnostic-Assessments}{Github link}\footnote{\url{https://github.com/kauroy1994/Large-Language-Models-for-Assisting-with-Mental-Health-Diagnostic-Assessments}} 
\paragraph*{\textbf{Institutional Review Board (IRB)}}
This study does not require approval from the Institutional Review Board (IRB). It involves using clinician-annotated social media posts, authorized for research purposes. The primary objective is to evaluate the effectiveness of LLMs that incorporate diagnostic criteria for major depressive disorder and general anxiety disorder for assisting with mental health assessments.

\section{Introduction}\label{sec:intro}
\begin{figure}[!htb]
    \centering
    \includegraphics[width=\linewidth,trim = 0cm 3cm 6cm 0cm, clip]{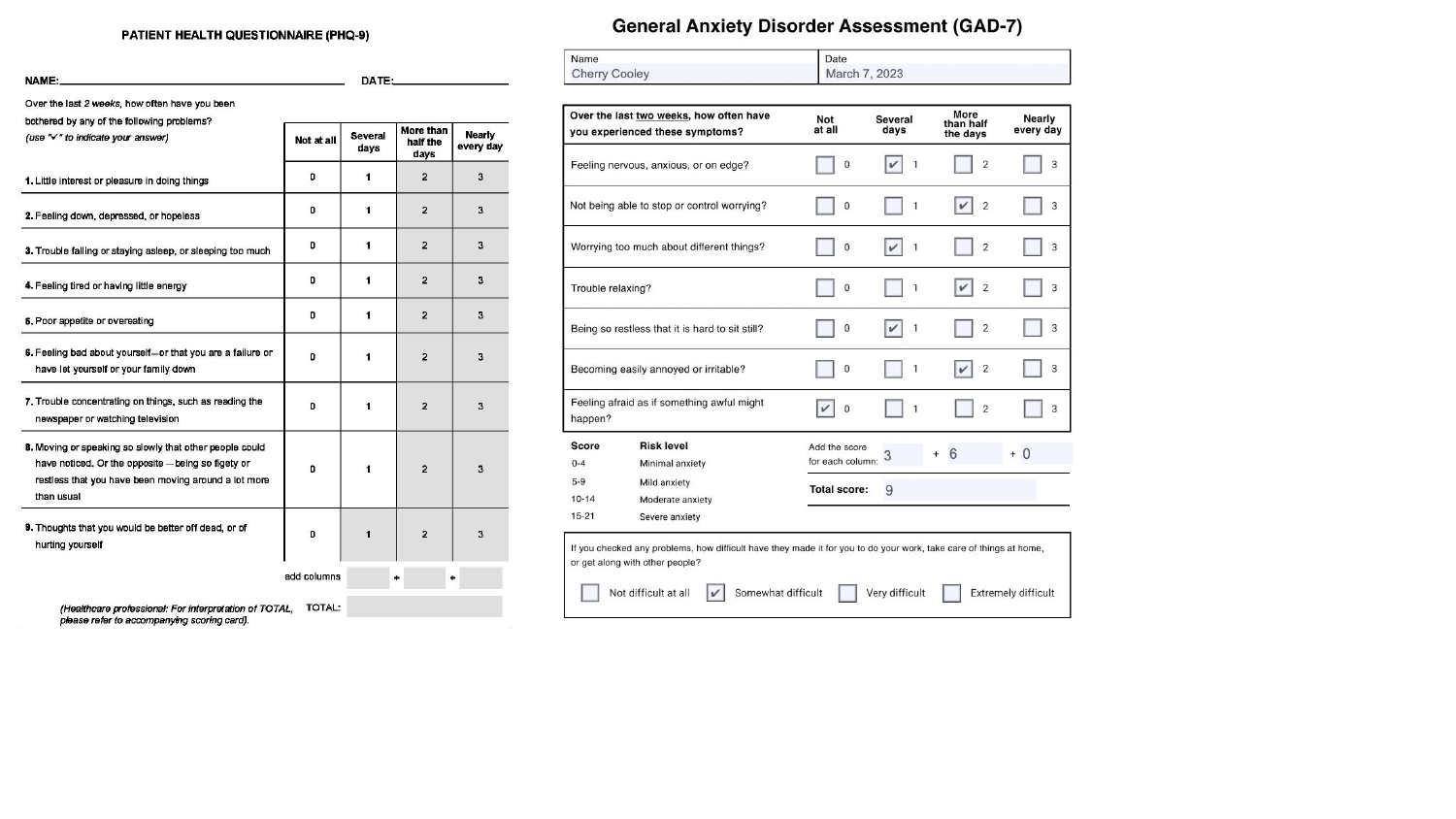}
    \caption{\textbf{Mental Health Diagnostic Assessment Questionnaires.} The Patient Health Questionnaire (PHQ)-9 for depression assessment and the Generalized Anxiety Disorder (GAD)-7 for anxiety assessment.}
    \label{fig:cprs}
\end{figure}

LLMs are large neural networks ($\geq \sim $7 billion weights and biases) designed to encode complex language patterns achieved by training on massive language-based datasets \cite{thirunavukarasu2023large}. Their remarkable success in a wide array of natural language processing tasks has led to the proliferation of LLM-based tools and applications across various industries \cite{rane2023chatgpt}. In healthcare, particularly in contexts involving natural language conversations, such as interactions between patients and clinicians, LLMs have piqued the interest of stakeholders as a potentially valuable tool to investigate for assisting with alleviating some of the burden on clinicians and the overall healthcare system \cite{lee2023benefits}. During patient-clinician interactions, clinicians employ standard diagnostic assessment processes for capturing a patient's state, such as the PHQ-9 for depression assessment and the GAD-7 for anxiety assessment \cite{ford2020use,johnson2019psychometric}. Figure \ref{fig:cprs} shows the PHQ-9 and GAD-7 questionnaires. 
To gainfully leverage LLMs for diagnostic assistance, it is necessary to provide mechanisms for guiding LLMs in closely following standardized clinical assessment procedures. There are two categories of methods available to enable this behavior:

(i) \textit{\textbf{Prompting LLMs}} - Modern LLMs stand out for their capacity to tailor responses based on user instructions or \textit{prompts} \cite{ouyang2022training}. However, LLMs are highly sensitive to the specific prompts used \cite{brown2020language}. Prompting techniques have continuously evolved to enhance the robustness of LLM responses, for example, by using Chain-of-Thought (CoT) prompting \cite{wei2022chain}. Prompting methods are broadly classified under three categories: (i) \textit{Naive prompting} - Providing direct instructions to the LLM in a prompt, (ii) \textit{Exemplar-based prompting} - Providing direct instructions along with few examples of the expected output, and (iii) \textit{Guidance-based prompting} - Exemplar-based prompting along with providing specific guidance on reasoning steps (for example by prompting the LLM to ``think'' step-by-step).

(ii) \textit{\textbf{Finetuning LLMs}} - Fine-tuning of LLMs involves adapting the model's behavior to closely align with the diagnostic procedures that clinicians follow, using fine-tuning algorithms such as supervised fine-tuning (SFT), reinforcement learning with human feedback (RLHF), and direct preference optimization (DPO) \cite{rafailov2024direct}. Fine-tuning LLMs is relatively more complex than prompting due to the need to curate high-quality data and appropriately formulate task-specific prompts or instructions during the fine-tuning process. 

In this work, we explore both approaches using a variety of proprietary and open-source models, namely - the Mentalllama and Llama models for finetuning, and the models GPT-3.5 and GPT-4o, llama-3.1-8b and mixtral-8x7b for prompting \cite{yang2024mentallama,touvron2023llama,jiang2024mixtral,brown2020language}. 
\subsection*{Related Work and Main Contribution}
Related work leveraging the PHQ-9 and GAD-7 questionnaires for diagnostic assistance for MDD and GAD, can be broadly categorized into: \textit{Scoring-based methods} - Scoring or ranking excerpts from text data (considered representative of patient-clinician interactions), based on relevance to symptoms presented in the PHQ-9 and GAD-7 questionnaires \cite{perez2023depresym}, \textit{Explainable AI (xAI)-based methods} -  that clinically ground BERT-based model outputs against the PHQ-9 and GAD-7 symptoms through surrogate modeling such as LIME and SHAP \cite{zirikly2022explaining}, and \textit{Text span identification and evidence summarization methods} - predicting and summarizing text spans over the data, and comparing against human annotated samples of highlighted text spans \cite{yates2024proceedings}. Our work most closely resembles the \textit{Text span identification and evidence summarization methods} category. However, our work differs in highly specialized steering of model outputs to provide information relevant to specific diagnostic criteria in the PHQ-9 and GAD-7 questionnaires across the prompting and fine-tuning methods employed in our experimentation. Additionally, we provide two significant contributions (1) a first-of-its-kind fine-tuned model specialized for diagnostic criteria assessment based on the llama model architecture, which we refer to as the $~\mathbf{\mathtt{Diagnostic}Llama}$, and (2) A comprehensive set of language model annotated synthetic data, evaluated for quality by expert humans for facilitating further research on LLM-powered diagnostic assessment.

\section{Methodology}\label{sec:methodology}
\subsection{MDD Diagnostic Assistance based on the PHQ-9}\label{subsec:phq-9}
\paragraph{\textbf{Ground Truth Dataset Creation}}
We start with the publicly available PRIMATE dataset, which consists of a collection of social media posts annotated for PHQ-9 relevant criteria  \cite{gupta2022learning}. Appendix \ref{app:primate} shows an example post and its annotation, specifically the post title, the post text, and the annotations indicating whether specific PHQ-9 symptoms are present in the post (using yes/no values). We chose this dataset as the authors provide preliminary experimental evidence on the effectiveness of using this dataset for guiding language models toward questionnaire-specific determination of diagnostic criteria. We first prompt GPT-4o to identify text spans in the posts corresponding to the PHQ-9 symptoms by providing an example of the expected output. Appendix \ref{app:gpt4o} shows an example of a prompt to GPT-4o. It is evident from this example how we are attempting to steer the model toward providing PHQ-9-specific diagnostic criteria. We then pass the model outputs to expert clinicians who provide us with a subset of GPT-4o annotated outputs that the clinicians agree with. This subset is available \href{https://huggingface.co/datasets/darssanle/GPT-4o-eval}{here}\footnote{\url{https://huggingface.co/datasets/darssanle/GPT-4o-eval}}. The clinicians are three anonymized experts from a non-profit \href{https://www.justdial.com/Bangalore/Dr-C-R-Chandrashekar-Samadhana-Counselling-Trust-Centre-Near-Subramanya-Temple-Mico-Layout-Bus-Stand-Bannerghatta-Road/080PXX80-XX80-170124231410-U5U4_BZDET}{institution} run by a retired professional from the National Institute of Mental Health, Neuroscience, and Allied Fields (NIMHANS), India\footnote{\url{https://www.justdial.com/Bangalore/Dr-C-R-Chandrashekar-Samadhana-Counselling-Trust-Centre-Near-Subramanya-Temple-Mico-Layout-Bus-Stand-Bannerghatta-Road/080PXX80-XX80-170124231410-U5U4_BZDET}}. The agreement score of $0.74$ was recorded among the annotators (measured using Cohen's Kappa). 
\subsubsection{\textbf{Prompting-based Methods}}\label{subsubsec:promptingPHQ9}
\paragraph{\textbf{Obtaining Proprietary Model Outputs for MDD Diagnostic Assistance based on the PHQ-9}}
Maintaining exactly the same prompt structure as shown in Appendix \ref{app:gpt4o}, we prompt the models GPT-3.5-Turbo and GPT-4o-mini to obtain annotations to a subset of the posts in the PRIMATE dataset. Our subset selection is randomized and limited by request costs and our available budget (see Section \ref{sec:ack} for funding information).

For evaluation of the outputs, we employ two methods, (i) \textit{hits@k based ranking} - We rank-order the text spans identified in the model output based on cosine similarity with the symptom, and then check if the identified text span occurs within the top k positions in the ground truth output, and (ii) \textit{Standard Classification Metrics} - We evaluate the accuracy, precision, recall and F1-score of the model outputs against the ground truth. Tables \ref{tab:prop_hits} and \ref{tab:prop_class} shows the evaluation results.
\begin{table}[!htb]
\centering
\caption{Evaluation of Proprietary LLMs for PHQ-9 Symptom Annotation of PRIMATE Posts Using hits@k.}
\begin{tabular}{lcc}
\toprule
     \textbf{Evaluation Metric} & \textbf{GPT-3.5-Turbo} & \textbf{GPT-4o-mini}  \\\midrule
     hits@1 & 87\% & \textbf{89}\%  \\
     hits@$<$5 & 98\% & \textbf{99}\% \\\bottomrule
\end{tabular}
\label{tab:prop_hits}
\end{table}

\begin{table}[!htb]
    \centering
    \caption{Evaluation of Proprietary LLMs for PHQ-9 Symptom Annotation of PRIMATE Posts Using Standard Classification Metrics.}
    \begin{tabular}{lcccc}
        \toprule
        \textbf{Method} & \textbf{Accuracy} & \textbf{Precision} & \textbf{Recall} & \textbf{F1-score} \\
        \midrule
        \textbf{GPT-3.5-Turbo} & 0.93 & 0.89 & 0.96  & 0.92 \\
        \textbf{GPT-4o-mini} & \textbf{0.94} & \textbf{0.96} & \textbf{0.98} & \textbf{0.92} \\

        \bottomrule
    \end{tabular}
    \label{tab:prop_class}
\end{table}
\subsubsection{\textbf{Obtaining Open-source Model Outputs for MDD Diagnostic Assistance based on the PHQ-9}.} 
Similar to the proprietary model case, we use the same prompt structure shown in Appendix \ref{app:gpt4o} and prompt the models llama3.1-8b and mixtral-8x7b to obtain annotations. Like the proprietary model(s) case, the subset selection is randomized and limited only by rate-limit costs. 

For evaluation, we use the same two methods defined in Section \ref{subsubsec:promptingPHQ9} using the ground truth dataset introduced in Section \ref{sec:methodology} (the hits@k and standard classification metrics). Tables \ref{tab:open_hits} and \ref{tab:open_class} show the results.
\begin{table}[!htb]
\centering
\caption{Evaluation of Open-source LLMs for PHQ-9 Symptom Annotation of PRIMATE Posts Using hits@k.}
\begin{tabular}{lcc}
\toprule
     \textbf{Evaluation Metric} & \textbf{llama3.1-8b} & \textbf{mixtral-8x7b}  \\\midrule
     hits@1 & 83\% & \textbf{92}\%  \\
     hits@$<$5 & 88\% & \textbf{99}\% \\\bottomrule
\end{tabular}
\label{tab:open_hits}
\end{table}

\begin{table}[!htb]
    \centering
    \caption{Evaluation of Open-source LLMs for PHQ-9 Symptom Annotation of PRIMATE Posts Using Standard Classification Metrics.}
    \begin{tabular}{lcccc}
        \toprule
        \textbf{Method} & \textbf{Accuracy} & \textbf{Precision} & \textbf{Recall} & \textbf{F1-score} \\
        \midrule
        \textbf{llama3.1-8b} & 0.84 & 0.86 & 0.78  & 0.82 \\
        \textbf{mixtral-8x7b} & \textbf{0.92} & \textbf{0.96} & \textbf{0.95} & \textbf{0.93} \\

        \bottomrule
    \end{tabular}
    \label{tab:open_class}
\end{table}
\subsubsection{\textbf{Fine-tuning-based Methods}}
\paragraph{\textbf{The MentalllaMa model}}
MentalllaMA is a model trained on 105K data samples of mental health instructions on social media posts. The samples are collected from 10 existing sources covering eight mental health analysis tasks, making MentalllaMA a suitable foundation model for the tasks covered in this study. The instructions used for training are a combination of expert-written and few-shot ChatGPT prompt outputs, further validating MentalllaMa as a viable candidate for testing adherence to diagnostic criteria by language models \cite{yang2024mentallama}. 

We perform experiments using MentaLLaMa on the ground truth dataset we introduce in Section \ref{sec:methodology} and report the results. The Prompt is provided in appendix Section \ref{app:mentalllama}. 

\paragraph{\textbf{The $~\mathbf{\mathtt{Diagnostic}Llama}$ model - Fine-tuning Mentalllama on the PRIMATE dataset using Hugging Face AutoTrain}}

Autorain is a no-code platform designed to simplify the process of training and fine-tuning language models on custom data\footnote{\url{https://huggingface.co/docs/autotrain/index}}. The full training specifications for training this model are available in appendix Section \ref{app:autotrain}. We refer to this model as $\mathbf{\mathtt{Diagnostic}Llama}$. Appendix section \ref{app:example} shows an example of an input (prompt) and output pair obtained using the $\mathbf{\mathtt{Diagnostic}Llama}$ model. The model space is available \href{https://huggingface.co/barca-boy/primate_autotrain_mental_llama}{here}\footnote{\url{https://huggingface.co/barca-boy/primate_autotrain_mental_llama}}. 

For evaluation of the outputs, we employ the same two methods as in Section \ref{subsubsec:promptingPHQ9}, i.e., (i) \textit{hits@k based ranking}, and (ii) \textit{Standard Classification Metrics} - the accuracy, precision, recall and F1-score of the model outputs against the ground truth. Tables \ref{tab:prop_hitsb} and \ref{tab:prop_classb} show the evaluation results.
\begin{table}[!htb]
\centering
\caption{Evaluation of MentalllaMa and $\mathbf{\mathtt{Diagnostic}Llama}$ for PHQ-9 Symptom Annotation of PRIMATE Posts Using hits@k.}
\begin{tabular}{lcc}
\toprule
     \textbf{Evaluation Metric} & \textbf{MentalllaMa} & \textbf{$\mathbf{\mathtt{Diagnostic}Llama}$}  \\\midrule
     hits@1 & - & \textbf{68.3}\%  \\
     hits@$<$5 & - & \textbf{76.2}\% \\\bottomrule
\end{tabular}
\label{tab:prop_hitsb}
\end{table}

\begin{table}[!htb]
    \centering
    \caption{Evaluation of MentalllaMa and $\mathbf{\mathtt{Diagnostic}Llama}$ for PHQ-9 Symptom Annotation of PRIMATE Posts Using Standard Classification Metrics.}
    \begin{tabular}{lcccc}
        \toprule
        \textbf{Method} & \textbf{Accuracy} & \textbf{Precision} & \textbf{Recall} & \textbf{F1-score} \\
        \midrule
        \textbf{MentalllaMa} & \textbf{0.82} & \textbf{0.83} & \textbf{0.63}  & \textbf{0.75} \\
        \textbf{$\mathbf{\mathtt{Diagnostic}Llama}$} & - & - & - & - \\

        \bottomrule
    \end{tabular}
    \label{tab:prop_classb}
\end{table}
\subsection{GAD Diagnostic Assistance based on the GAD-7}
\paragraph{\textbf{Ground Truth Dataset Creation}}
Once again, we start with the publicly available PRIMATE dataset. We then prompt GPT-4o to identify text spans in the posts corresponding to the GAD-7 symptoms by providing an example of the expected output. Appendix \ref{app:gpt4o} shows an example of a prompt to GPT-4o. This example clarifies how we attempt to steer the model toward providing GAD-7-specific diagnostic criteria. Similar to the PHQ-9 case, we then pass the model outputs to expert clinicians who provide us with a subset of GPT-4o annotated outputs that the clinicians agree with. This subset is available \href{https://huggingface.co/datasets/darssanle/GPT-4o-GAD-7}{here}\footnote{\url{https://huggingface.co/datasets/darssanle/GPT-4o-GAD-7}}. The clinicians are the same three anonymized experts from the non-profit mentioned in Section \ref{subsec:phq-9}. The agreement score of $0.72$ was recorded among the annotators (measured using Cohen's Kappa). 
\subsubsection{\textbf{Prompting-based Methods}}
\paragraph{\textbf{Obtaining Proprietary Model Outputs for MDD Diagnostic Assistance based on the GAD-7}}
Maintaining exactly the same prompt structure as shown in Appendix \ref{app:gpt4o}, we prompt the models GPT-3.5-Turbo and GPT-4o-mini to obtain annotations to a subset of the posts in the PRIMATE dataset, but this time geared towards responses to the GAD-7 symptoms. Our subset selection is randomized and limited by request costs and our available budget (see Section \ref{sec:ack} for funding information).

For evaluation of the outputs, we employ the same two methods as in the PHQ-9 case, i.e., (i) \textit{hits@k based ranking}, and (ii) \textit{Standard Classification Metrics} - the accuracy, precision, recall and F1-score of the model outputs against the ground truth. Tables \ref{tab:prop_hits_2} and \ref{tab:prop_class_2} show the evaluation results.
\begin{table}[!htb]
\centering
\caption{Evaluation of Proprietary LLMs for GAD-7 Symptom Annotation of PRIMATE Posts Using hits@k.}
\begin{tabular}{lcc}
\toprule
     \textbf{Evaluation Metric} & \textbf{GPT-3.5-Turbo} & \textbf{GPT-4o-mini}  \\\midrule
     hits@1 & 88\% & \textbf{89}\%  \\
     hits@$<$5 & 98\% & \textbf{98}\% \\\bottomrule
\end{tabular}
\label{tab:prop_hits_2}
\end{table}

\begin{table}[!htb]
    \centering
    \caption{Evaluation of Proprietary LLMs for GAD-7 Symptom Annotation of PRIMATE Posts Using Standard Classification Metrics.}
    \begin{tabular}{lcccc}
        \toprule
        \textbf{Method} & \textbf{Accuracy} & \textbf{Precision} & \textbf{Recall} & \textbf{F1-score} \\
        \midrule
        \textbf{GPT-3.5-Turbo} & 0.95 & 0.9 & 0.95  & 0.91 \\
        \textbf{GPT-4o-mini} & 0.93 & 0.97 & 0.91 & 0.92 \\

        \bottomrule
    \end{tabular}
    \label{tab:prop_class_2}
\end{table}
\subsubsection{\textbf{Obtaining Open-source Model Outputs for MDD Diagnostic Assistance based on the GAD-7}.} 
Like the PHQ-9, we use the same prompt structure shown in Appendix \ref{app:gpt4o} and prompt the models llama3.1-8b and mixtral-8x7b to obtain annotations to the GAD-7 symptoms. As before, the subset selection is randomized and limited only by rate-limit costs. 

For evaluation, we use the same two methods defined in Section \ref{subsubsec:promptingPHQ9} using the ground truth dataset introduced in Section \ref{sec:methodology} (the hits@k and standard classification metrics). Tables \ref{tab:open_hitsb} and \ref{tab:open_classb} show the results.
\begin{table}[!htb]
\centering
\caption{Evaluation of Open-source LLMs for GAD-7 Symptom Annotation of PRIMATE Posts Using hits@k.}
\begin{tabular}{lcc}
\toprule
     \textbf{Evaluation Metric} & \textbf{llama3.1-8b} & \textbf{mixtral-8x7b}  \\\midrule
     hits@1 & 83\% & \textbf{92}\%  \\
     hits@$<$5 & 88\% & \textbf{99}\% \\\bottomrule
\end{tabular}
\label{tab:open_hitsb}
\end{table}

\begin{table}[!htb]
    \centering
    \caption{Evaluation of Open-source LLMs for GAD-7 Symptom Annotation of PRIMATE Posts Using Standard Classification Metrics.}
    \begin{tabular}{lcccc}
        \toprule
        \textbf{Method} & \textbf{Accuracy} & \textbf{Precision} & \textbf{Recall} & \textbf{F1-score} \\
        \midrule
        \textbf{llama3.1-8b} & 0.84 & 0.86 & 0.78  & 0.82 \\
        \textbf{mixtral-8x7b} & \textbf{0.92} & \textbf{0.96} & \textbf{0.95} & \textbf{0.93} \\

        \bottomrule
    \end{tabular}
    \label{tab:open_classb}
\end{table}
\subsection{A Note on older LLMs and Classification-based Approaches}
\paragraph{Older Autoregressive LLMs}
The previous sections have covered the best-performing LLMs. However, we have performed experiments on older LLMs such as Llama2 and Mistral, and we provide these results in Table \ref{tab:llms_old} \cite{touvron2023llama,jiang2023mistral}.
\begin{table}[!htb]
    \centering
    \caption{Evaluation of Llama2-7b-chat and Mistral-Instruct for PHQ-9 Symptom Annotations of the PRIMATE Posts Using F1 scores.}
    \begin{tabular}{lc}
        \toprule
        \textbf{Method} & \textbf{F1-score} \\
        \midrule
        \textbf{llama2-7b-chat} & 0.663 \\
        \textbf{mistral-instruct} & \textbf{0.655} \\
        \bottomrule
    \end{tabular}
    \label{tab:llms_old}
\end{table}
\paragraph{Older pretrained language models}
Several classification-based approaches have been used to classify posts into labels corresponding to diagnostic criteria on questionnaires as an alternative to generative models \cite{dalal2023cross,gupta2022learning}. Although this work focuses on modern LLMs, we also perform experiments in the classification setting using the older pretrained models - BERT, MentalBERT, and MentalRoBERTa \cite{devlin2018bert,ji2022mentalbert}. Table \ref{tab:plms} shows the results\footnote{For completeness, we also show results of traditional machine learning-based classification methods in appendix Section \ref{app:ml}}.
\begin{table}[!htb]
    \centering
    \caption{Evaluation of BERT, MentalBERT, and MentalRoBERTa for PHQ-9 Symptom Annotations of the PRIMATE Posts Using F1 scores.}
    \begin{tabular}{lc}
        \toprule
        \textbf{Method} & \textbf{F1-score} \\
        \midrule
        \textbf{BERT} & 0.69 \\
        \textbf{MentalBERT} & \textbf{0.71} \\
        \textbf{MentalRoBERTa} & 0.48 \\
        \bottomrule
    \end{tabular}
    \label{tab:plms}
\end{table}
\subsection{Model and Data Artifact Details}\label{subsec:artifacts}
As part of this study, we release several software artifacts, including \textbf{one model} - the $\mathbf{\mathtt{Diagnostic}Llama}$ model (available \href{https://huggingface.co/barca-boy/primate_autotrain_mental_llama}{here}\footnote{\url{https://huggingface.co/barca-boy/primate_autotrain_mental_llama}}), and multiple annotated datasets that contain diagnostic symptom predictions along with text-span highlights categorized into: 

\textbf{(a) PHQ-9-based Annotations}, namely - (i) GPT-3.5-PHQ-9 annotations (ii) GPT-4o\_mini-PHQ-9 annotations, (iii) GPT-4o-PHQ-9 annotations, (iv) llama3.1\_8b-PHQ-9 annotations, (v) mixtral-8x7b-PHQ-9 annotations, and 

\textbf{(b) containing GAD-7 based annotations} (i) GPT-3.5-GAD-7 annotations (ii) GPT-4o\_mini-GAD-7 annotations, (iii) GPT-4o-GAD-7 annotations, (iv) llama3.1\_8b-GAD-7 annotations, (v) mixtral-8x7b-GAD-7 annotations.

The datasets are all available at this \href{https://huggingface.co/collections/darssanle/mhd-datasets-669628ee2d25bd04e99dc3bf}{link}\footnote{\url{https://huggingface.co/collections/darssanle/mhd-datasets-669628ee2d25bd04e99dc3bf}}. The dataset statistics are available as part of the dataset cards in the links provided. The dataset cards also show the details of the prompting structure used to generate the LLM outputs. We have also consolidated all the links to the model and data artifacts in this \href{https://github.com/kauroy1994/Large-Language-Models-for-Assisting-with-Mental-Health-Diagnostic-Assessments}{Github repository}\footnote{\url{https://github.com/kauroy1994/Large-Language-Models-for-Assisting-with-Mental-Health-Diagnostic-Assessments}}. Table \ref{tab:data_stats} provides a summary.
\begin{table}[!htb]
\centering
\caption{Dataset Statistics (number of posts) for All the Datasets in Section \ref{subsec:artifacts}}
\begin{tabular}{lc}
\toprule
     \textbf{Dataset} & \textbf{Number of Posts} \\\midrule
     GPT-3.5-PHQ-9 & 339  \\
     GPT-4o\_mini-PHQ-9 & 102 \\
     GPT-4o-PHQ-9 & 40\\
     llama3.1\_8b-PHQ-9 & 155\\
     mixtral-8x7b-PHQ-9 & 97\\
     GPT-4o\_mini-GAD-7 & 51 \\
     GPT-4o-GAD-7 & 17\\
     llama3.1\_8b-GAD-7 & 124\\
     mixtral-8x7b-GAD-7 & 109\\\bottomrule
     \textbf{Total} & 1034\\\bottomrule
\end{tabular}
\label{tab:data_stats}
\end{table}
\section{Results}
\subsection{PHQ-9 Results}
From Tables \ref{tab:prop_hits}, \ref{tab:prop_class}, \ref{tab:open_hits}, and \ref{tab:open_class}, we see that both the proprietary and open-source LLMs approach human annotation quality, and Tables \ref{tab:prop_hitsb} and \ref{tab:prop_classb} show that fine-tuning LLMs for diagnostic assistance shows promising results. However, fine-tuning LLMs has turned out to be highly challenging and needs considerable resources and hyperparameter tuning to get right. The entries for MentalllaMa are blank in the tables because the MentalllaMa model reiterates the input verbatim, as seen in Section \ref{app:mentalllama}. This further shows the difficulty of adequately leveraging fine-tuned models to achieve good results in highly specialized tasks such as diagnostic assistance. Still, the preliminary results on the PHQ-9 task demonstrate that this can be done with a good bit of trial and error on the fine-tuning configurations. It is essential to be able to deploy specialized models fine-tuned/trained on custom data in safety-constrained and privacy-critical settings. 

Interestingly, Table \ref{tab:llms_old} shows significant performance gaps between the older and newer LLMs (open-source and proprietary models). We also find from Table \ref{tab:plms} that older pretrained language models (that are not autoregressive), perform as well as older LLMs. We also see again that fine-tuning in the case of pretrained LLMs does not lead to much change in performance and sometimes leads to bad performance (e.g., MentalRoBERTa), further evidencing the significant challenge with fine-tuning language models for specialized tasks such as mental health diagnostic assistance.

\subsection{GAD-7 Results}
For the GAD-7 results, from Tables \ref{tab:prop_hitsb}, \ref{tab:prop_classb}, \ref{tab:open_hitsb}, and \ref{tab:open_classb} we see a similar trend as the PHQ-9 case, i.e., both proprietary and open-source LLMs approach human annotations quality.

Among the proprietary models, we find GPT-4o\_mini to be the best performing, and mixtral-8x7b-GAD-7 among the open-source models. However, there are no significant differences between the different LLMs, including both proprietary and open-source LLMs.

\section{Conclusion and Future Work}
Previous efforts in utilizing Large Language Models (LLMs) for mental health assistance have primarily focused on conversational data or diagnostic assessment as a classification problem. However, these initiatives lack the precision and guidance necessary for effective assessment with robust explanations (reasoning), and response generation, based on established questionnaires. This gap is significant because standardized assessment tools, such as the PHQ-9 for major depressive disorder and the GAD-7 for general anxiety disorder, are indispensable for accurate and effective treatment planning in mental healthcare. Our research addresses this gap by specifically targeting these assessment procedures and developing prompting strategies to guide LLMs toward crafting clinician-friendly responses with explanations using assessment and reasoning prompts.

Our findings reveal that while LLMs struggle to effectively utilize questionnaire information in prompts to provide assessments resembling those of clinicians in the zero-shot setting, their performance significantly improves in the few-shot setting (both in fine-tuning and few-shot prompting regime), nearly matching the assessments of expert clinicians. However, despite this improvement, LLMs still do not reason in the same manner as clinicians when arriving at assessments, matching clinician reasoning only a fraction of the time, as evidenced by the sizes of the ground truth datasets for which a high expert agreement score is obtained. This underscores the need for further scrutiny in the integration of LLMs, along with prompting methods incorporating diagnostic assessment criteria, before they can be reliably utilized in mental healthcare assistance. Moreover, our work introduces several novel assessment LLM and instruction-tuning datasets, offering a valuable resource for future research aimed at understanding and enhancing the effectiveness of LLMs in assisting with assessments within mental healthcare settings. This contribution holds promise for advancing the capabilities of LLMs in mental health support, potentially alleviating the strain on healthcare systems caused by a shortage of care providers and an increasing number of patients. 

\paragraph{\textbf{Future Work}} We are working on integrating the models studied in this work into a clinician-facing app, and extending the $\mathbf{\mathtt{Diagnostic}Llama}$ model to include GAD-7, and expanding all datasets in Section \ref{subsec:artifacts} to match the original PRIMATE dataset. We are also expanding our datasets and results to include more GAD-7-based results and non-linearly structured questionnaires (example flowcharts) such as the CSSRS \cite{roy2023process}\footnote{app demo link: \url{https://www.youtube.com/watch?v=VpJYyb7brRs&list=PLqJzTtkUiq577Rc1HpX4iE1_ntNeuppzA&index=22}}. Finally, we are also working to incorporate additional constraints, such as restricted terminology (e.g., non-toxic terminology), by paraphrasing the LLM outputs \cite{tsakalidis2022overview,roy2023proknow}. All future updates will be released on the GitHub repository.
\section*{Acknowledgements}\label{sec:ack}
This research is partially supported by \href{https://www.nsf.gov/awardsearch/showAward?AWD_ID=2335967}{NSF Award 2335967 EAGER: Knowledge-guided neurosymbolic AI with guardrails for safe virtual health assistants}\footnote{\url{https://www.nsf.gov/awardsearch/showAward?AWD_ID=2335967}} \cite{sheth2021knowledge,sheth2022process,sheth2023neurosymbolic,sheth2024neurosymbolic,sheth2024civilizing,sheth2024neurosymbolicb}. The views expressed here are those of the authors, not those of the sponsors.
\bibliographystyle{unsrt}
\bibliography{acmart}
\centering\section*{Appendix}
\appendix

\section{Dataset Examples}
\subsection{Primate Data Example}\label{app:primate}
\begin{lstlisting}
{
    "post_title": "I don't feel original anymore.",
    "post_text": "When I was in high school a few years back, I was one of the highest competitors in my school. I joined the high school band in freshman year and by senior year I became one of the best in my section. My academics were always straight and I exercised daily. Senior year I enlisted in the military and now I believe it was one of my worst decisions in life. Before I went to boot camp I was motivated, a patriot and believed that the elite joined the military. In senior year I never applied for any scholarships and I was offered one but turned it down because I already signed the papers. I thought I set myself up for success. Now I believe I was dead wrong for joining. The only benefit I see so far after a year and a half of service is that I'm trying to set myself up financially before I get out and hopefully attend college. It sounds like a plan but I feel no happiness from what I do at all. I convinced myself there's no honor in it anymore, it's just another job. I don't exercise by myself anymore. I feel like I'm not progressing anywhere in life being in service. I'm just a body and if I wasn't here doing what I'm doing, there'd just be somebody else doing the exact same. I'm replaceable. That's the mindset the military gave me. I look forward to going back home in 6 months for vacation and that's the only thing I've been looking forward to since I've been stationed. After that, the only thing I have my eyes on are getting out of service, going home, being closer to my family again. There's nothing here that satisfies me and I hate it. I feel like I've tried everything to be happy here but it seems impossible. I wish somebody could help.",
    "annotations": [
      [
        "Feeling-bad-about-yourself-or-that-you-are-a-failure-or-have-let-yourself-or-your-family-down",
        "yes"
      ],
      [
        "Feeling-down-depressed-or-hopeless",
        "no"
      ],
      [
        "Feeling-tired-or-having-little-energy",
        "yes"
      ],
      [
        "Little-interest-or-pleasure-in-doing",
        "yes"
      ],
      [
        "Moving-or-speaking-so-slowly-that-other-people-could-have-noticed-Or-the-opposite-being-so-fidgety-or-restless-that-you-have-been-moving-around-a-lot-more-than-usual",
        "no"
      ],
      [
        "Poor-appetite-or-overeating",
        "no"
      ],
      [
        "Thoughts-that-you-would-be-better-off-dead-or-of-hurting-yourself-in-some-way",
        "no"
      ],
      [
        "Trouble-concentrating-on-things-such-as-reading-the-newspaper-or-watching-television",
        "no"
      ],
      [
        "Trouble-falling-or-staying-asleep-or-sleeping-too-much",
        "no"
      ]
    ]
}
\end{lstlisting}
\subsection{GPT-4o Prompt Example}\label{app:gpt4o}
\begin{lstlisting}
""" When given the below JSON formatted file content, I need you to give me the specific sentences from the text that exhibit a set of symptoms. Below is an example of INPUT and OUTPUT. Keep JSON Formatting for output:

{
    "post_title": "I don't feel original anymore.",
    "post_text": "When I was in high school a few years back, I was one of the highest competitors in my school. I joined the high school band in freshman year and by senior year I became one of the best in my section. My academics were always straight and I exercised daily. Senior year I enlisted in the military and now I believe it was one of my worst decisions in life. Before I went to boot camp I was motivated, a patriot and believed that the elite joined the military. In senior year I never applied for any scholarships and I was offered one but turned it down because I already signed the papers. I thought I set myself up for success. Now I believe I was dead wrong for joining. The only benefit I see so far after a year and a half of service is that I'm trying to set myself up financially before I get out and hopefully attend college. It sounds like a plan but I feel no happiness from what I do at all. I convinced myself there's no honor in it anymore, it's just another job. I don't exercise by myself anymore. I feel like I'm not progressing anywhere in life being in service. I'm just a body and if I wasn't here doing what I'm doing, there'd just be somebody else doing the exact same. I'm replaceable. That's the mindset the military gave me. I look forward to going back home in 6 months for vacation and that's the only thing I've been looking forward to since I've been stationed. After that, the only thing I have my eyes on are getting out of service, going home, being closer to my family again. There's nothing here that satisfies me and I hate it. I feel like I've tried everything to be happy here but it seems impossible. I wish somebody could help.",
    "annotations": [
      [
        "Feeling-bad-about-yourself-or-that-you-are-a-failure-or-have-let-yourself-or-your-family-down",
        "yes"
      ],
      [
        "Feeling-down-depressed-or-hopeless",
        "no"
      ],
      [
        "Feeling-tired-or-having-little-energy",
        "yes"
      ],
      [
        "Little-interest-or-pleasure-in-doing",
        "yes"
      ],
      [
        "Moving-or-speaking-so-slowly-that-other-people-could-have-noticed-Or-the-opposite-being-so-fidgety-or-restless-that-you-have-been-moving-around-a-lot-more-than-usual",
        "no"
      ],
      [
        "Poor-appetite-or-overeating",
        "no"
      ],
      [
        "Thoughts-that-you-would-be-better-off-dead-or-of-hurting-yourself-in-some-way",
        "no"
      ],
      [
        "Trouble-concentrating-on-things-such-as-reading-the-newspaper-or-watching-television",
        "no"
      ],
      [
        "Trouble-falling-or-staying-asleep-or-sleeping-too-much",
        "no"
      ]
    ]
}

And this is an example expected output format:

{
    "post_title": "I don't feel original anymore.",
    "post_text": "When I was in high school a few years back, I was one of the highest competitors in my school. I joined the high school band in freshman year and by senior year I became one of the best in my section. My academics were always straight, and I exercised daily. Senior year I enlisted in the military, and now I believe it was one of my worst decisions in life. Before I went to boot camp I was motivated, a patriot and believed that the elite joined the military. In senior year I never applied for any scholarships and I was offered one but turned it down because I already signed the papers. I thought I set myself up for success. Now I believe I was dead wrong for joining. The only benefit I see so far after a year and a half of service is that I'm trying to set myself up financially before I get out and hopefully attend college. It sounds like a plan but I feel no happiness from what I do at all. I convinced myself there's no honor in it anymore; it's just another job. I don't exercise by myself anymore. I feel like I'm not progressing anywhere in life being in service. I'm just a body, and if I wasn't here doing what I'm doing, there'd just be somebody else doing the exact same. I'm replaceable. That's the mindset the military gave me. I look forward to going back home in 6 months for vacation, and that's the only thing I've been looking forward to since I've been stationed. After that, the only thing I have my eyes on is getting out of service, going home, being closer to my family again. There's nothing here that satisfies me, and I hate it. I feel like I've tried everything to be happy here but it seems impossible. I wish somebody could help.",
    "annotations": {
      "Feeling-bad-about-yourself-or-that-you-are-a-failure-or-have-let-yourself-or-your-family-down": [
        "I thought I set myself up for success. Now I believe I was dead wrong for joining."
      ],
      "Feeling-down-depressed-or-hopeless": [],
      "Feeling-tired-or-having-little-energy": [
        "I feel like I'm not progressing anywhere in life being in service."
      ],
      "Little-interest-or-pleasure-in-doing": [
        "There's nothing here that satisfies me, and I hate it."
      ],
      "Moving-or-speaking-so-slowly-that-other-people-could-have-noticed-Or-the-opposite-being-so-fidgety-or-restless-that-you-have-been-moving-around-a-lot-more-than-usual": [],
      "Poor-appetite-or-overeating": [],
      "Thoughts-that-you-would-be-better-off-dead-or-of-hurting-yourself-in-some-way": [],
      "Trouble-concentrating-on-things-such-as-reading-the-newspaper-or-watching-television": [],
      "Trouble-falling-or-staying-asleep-or-sleeping-too-much": []
    }
  },

May I proceed with the rest of the INPUTS? """
\end{lstlisting}
\section{MentalllaMa}\label{app:mentalllama_details}
\subsection{Example Input and Output}\label{app:mentalllama}
\textbf{Input}

\begin{lstlisting}
    ### INSTRUCTION: 


 For a given user post sentence, does it show signs of the symptom. Answer in binary "yes" or
"no", for every symptom. The symptoms are as follows:
[Little interest or pleasure in doing things,

Feeling down, depressed, or hopeless,

Trouble falling or staying asleep, or sleeping too much,
Feeling tired or having little energy,

Poor appetite or overeating,

Feeling bad about yourself or that you are a failure or
have let yourself or your family down,

Trouble concentrating on things, such as reading the
newspaper or watching television,

Moving or speaking so slowly that other people could
have noticed. Or the opposite being so figety or
restless that you have been moving around a lot more
than usual,

Thoughts that you would be better off dead, or of
hurting yourself] .

Give the output as a python list of lists, with the list containing the symptom and the
answer, separated by a comma.



 ### INPUT: 
Hello. I (23F) have tried to make a post to explain what's going on with me, but I just end up crying, and everything I write is incoherent, so. Sorry if there's not enough context?

I'm having a really difficult time right now. I can't really focus on work, and I don't get in as many hours as I should. I feel like I am letting people down. 

Existing is exhausting, and all I can do is waste time on my phone, because if I'm on my phone I don't have to think and time passes more quickly. 
I'm tired of spending hours on my phone instead of doing functional-people things, I'm just putting everything off and setting myself up for failure. But doing anything productive feels too daunting, and I don't know how to start? 

I want to develop healthy habits - spend less time on my phone, maybe go outside more often, go on walks, stick to my work schedule, find hobbies. Simple things that I know would improve my mental health.
But even though I want those things, I feel like I'm not motivated enough to make a change. It's always easier to stay in my comfort zone, to distract myself. My brain says I need to change, but I feel like I don't want it bad enough to actually try. 

I can't keep living like this, though. 

I've tried looking for advice online, but everything I see involves:
- Therapy, which I don't have money for. 
- Reaching out, but I have very few people I'm close with, and I don't feel comfortable talking about my mental health to any of them. 
- Self-help books and stuff, which just...does not work on me.
- Hobbies, but I don't really have any of those anymore? I've been depressed for so long, I've lost interest in the ones I used to have as a kid, and never developed new ones. 

So I guess I'm wondering/hoping you all might have some advice on how to get a bit better, something that doesn't involve the things I just mentioned ? 

I don't know. I wish wanting to get better was enough to spur me into action. Sorry if what I'm asking for doesn't exist, or if I wasn't very clear.
Thanks for reading
\end{lstlisting}

\textbf{Output}
\begin{lstlisting}
    ### INSTRUCTION: 


 For a given user post sentence, does it show signs of the symptom. Answer in binary "yes" or
"no", for every symptom. The symptoms are as follows:
[Little interest or pleasure in doing things,

Feeling down, depressed, or hopeless,

Trouble falling or staying asleep, or sleeping too much,
Feeling tired or having little energy,

Poor appetite or overeating,

Feeling bad about yourself or that you are a failure or
have let yourself or your family down,

Trouble concentrating on things, such as reading the
newspaper or watching television,

Moving or speaking so slowly that other people could
have noticed. Or the opposite being so figety or
restless that you have been moving around a lot more
than usual,

Thoughts that you would be better off dead, or of
hurting yourself] .

Give the output as a python list of lists, with the list containing the symptom and the
answer, separated by a comma.



 ### INPUT: 
Hello. I (23F) have tried to make a post to explain what's going on with me, but I just end up crying, and everything I write is incoherent, so. Sorry if there's not enough context?

I'm having a really difficult time right now. I can't really focus on work, and I don't get in as many hours as I should. I feel like I am letting people down. 

Existing is exhausting, and all I can do is waste time on my phone, because if I'm on my phone I don't have to think and time passes more quickly. 
I'm tired of spending hours on my phone instead of doing functional-people things, I'm just putting everything off and setting myself up for failure. But doing anything productive feels too daunting, and I don't know how to start? 

I want to develop healthy habits - spend less time on my phone, maybe go outside more often, go on walks, stick to my work schedule, find hobbies. Simple things that I know would improve my mental health.
But even though I want those things, I feel like I'm not motivated enough to make a change. It's always easier to stay in my comfort zone, to distract myself. My brain says I need to change, but I feel like I don't want it bad enough to actually try. 

I can't keep living like this, though. 

I've tried looking for advice online, but everything I see involves:
- Therapy, which I don't have money for. 
- Reaching out, but I have very few people I'm close with, and I don't feel comfortable talking about my mental health to any of them. 
- Self-help books and stuff, which just...does not work on me.
- Hobbies, but I don't really have any of those anymore? I've been depressed for so long, I've lost interest in the ones I used to have as a kid, and never developed new ones. 

So I guess I'm wondering/hoping you all might have some advice on how to get a bit better, something that doesn't involve the things I just mentioned ? 

I don't know. I wish wanting to get better was enough to spur me into action. Sorry if what I'm asking for doesn't exist, or if I wasn't very clear.
Thanks for reading
\end{lstlisting}

\section{Autotraining $\mathtt{Diagnostic}Llama$}\label{app:autotrain}
\textbf{Model details, input format, sample inferences}
\subsection{Example Input and Output}\label{app:example}
\textbf{Input}

\begin{lstlisting}
    ### INSTRUCTION: 


 For a given user post sentence, does it show signs of the symptom. Answer in binary "yes" or
"no", for every symptom. The symptoms are as follows:
[Little interest or pleasure in doing things,

Feeling down, depressed, or hopeless,

Trouble falling or staying asleep, or sleeping too much,
Feeling tired or having little energy,

Poor appetite or overeating,

Feeling bad about yourself or that you are a failure or
have let yourself or your family down,

Trouble concentrating on things, such as reading the
newspaper or watching television,

Moving or speaking so slowly that other people could
have noticed. Or the opposite being so figety or
restless that you have been moving around a lot more
than usual,

Thoughts that you would be better off dead, or of
hurting yourself] .

Give the output as a python list of lists, with the list containing the symptom and the
answer, separated by a comma.



 ### INPUT: 
Hello. I (23F) have tried to make a post to explain what's going on with me, but I just end up crying, and everything I write is incoherent, so. Sorry if there's not enough context?

I'm having a really difficult time right now. I can't really focus on work, and I don't get in as many hours as I should. I feel like I am letting people down. 

Existing is exhausting, and all I can do is waste time on my phone, because if I'm on my phone I don't have to think and time passes more quickly. 
I'm tired of spending hours on my phone instead of doing functional-people things, I'm just putting everything off and setting myself up for failure. But doing anything productive feels too daunting, and I don't know how to start? 

I want to develop healthy habits - spend less time on my phone, maybe go outside more often, go on walks, stick to my work schedule, find hobbies. Simple things that I know would improve my mental health.
But even though I want those things, I feel like I'm not motivated enough to make a change. It's always easier to stay in my comfort zone, to distract myself. My brain says I need to change, but I feel like I don't want it bad enough to actually try. 

I can't keep living like this, though. 

I've tried looking for advice online, but everything I see involves:
- Therapy, which I don't have money for. 
- Reaching out, but I have very few people I'm close with, and I don't feel comfortable talking about my mental health to any of them. 
- Self-help books and stuff, which just...does not work on me.
- Hobbies, but I don't really have any of those anymore? I've been depressed for so long, I've lost interest in the ones I used to have as a kid, and never developed new ones. 

So I guess I'm wondering/hoping you all might have some advice on how to get a bit better, something that doesn't involve the things I just mentioned ? 

I don't know. I wish wanting to get better was enough to spur me into action. Sorry if what I'm asking for doesn't exist, or if I wasn't very clear.
Thanks for reading
\end{lstlisting}

\textbf{Output}

\begin{lstlisting}
    [['Feeling-bad-about-yourself-or-that-you-are-a-failure-or-have-let-yourself-or-your-family-down', 'yes'], ['Feeling-down-depressed-or-hopeless', 'yes'], ['Feeling-tired-or-having-little-energy', 'yes'], ['Little-interest-or-pleasure-in-doing ', 'yes'], ['Moving-or-speaking-so-slowly-that-other-people-could-have-noticed-Or-the-opposite-being-so-fidgety-or-restless-that-you-have-been-moving-around-a-lot-more-than-usual', 'no'], ['Poor-appetite-or-overeating', 'no'], ['Thoughts-that-you-would-be-better-off-dead-or-of-hurting-yourself-in-some-way', 'no'], ['Trouble-concentrating-on-things-such-as-reading-the-newspaper-or-watching-television', 'yes'], ['Trouble-falling-or-staying-asleep-or-sleeping-too-much', 'no']]
\end{lstlisting}
\section{Traditional Machine Learning-based Approaches}\label{app:ml}
\begin{table}[!htb]
    \centering
    \caption{Evaluation of Traditional ML-based methods for PHQ-9 Symptom Annotations of the PRIMATE Posts Using F1 scores.}
    \begin{tabular}{lc}
        \toprule
        \textbf{Method} & \textbf{F1-score} \\
        \midrule
        \textbf{Logistic Regression} & 0.49 \\
        \textbf{Random Forest} & 0.38 \\
        \textbf{XGBoost} & 0.65 \\
        \bottomrule
    \end{tabular}
    \label{tab:ml}
\end{table}
\end{document}